\documentclass{article}
\usepackage{spconf,amsmath,graphicx,hyperref}
\usepackage{amssymb,booktabs,xcolor,multirow,tikz}

\DeclareMathOperator*{\argmax}{arg\,max}
\DeclareMathOperator{\clip}{clip}
\DeclareMathOperator{\mean}{mean}
\DeclareMathOperator{\std}{std}

\title{MVAgent: Multi-Agent Video Generation via Consistent Condition Construction and Shot-Level Policy Optimization}
\name{\begin{tabular}{@{}c@{}}
Xiangyu Kong$^{1,2,\dagger}$,
Wenjie Zhou$^{2,\dagger}$,
Fengping Tian$^{2}$,
Lihua Fang$^{2}$,
Haoqin Sun$^{2}$,\\
Chenyang Lyu$^{2,*}$,
Longyue Wang$^{2}$,
Weihua Luo$^{2}$
\thanks{$^*$Corresponding author. $^\dagger$Equal contribution.}
\end{tabular}}

\address{
$^{1}$University of Exeter, United Kingdom\\
$^{2}$Alibaba Group, China\\
}

\begin{document}
\ninept
\maketitle
\begin{abstract}

Multi-shot agentic video generation requires consistent character appearance, stable spatial layout across camera angles, and continuous character state between shots. When every shot is a separate request to a frozen generator, repeated text does not determine appearance, layout or state. We therefore recast the problem as condition construction and present MVAgent, a multi-agent pipeline whose agents collaborate through typed conditioning inputs. Because an environment image shows one viewpoint, a Spatial Grounding agent samples views from generated camera-traversal clips and anchors each shot to the view matching its framing. As generated shots drift from the plan, an Observer records how each shot ends in a continuity memory, from which a Transition agent builds character action and spatial references for the next shot. An Orchestrator composes these inputs into each request. Since a request reveals its effect only after rendering, we train it by agentic reinforcement learning with Trunk-GDPO, which compares rendered candidates at every shot rather than once per video and continues the best as the trunk. With generator and judges frozen, MVAgent attains the highest cross-shot consistency and narrative-planning quality among the compared methods on ViMax-Bench and is preferred over the strongest agentic baseline in human evaluation.
\end{abstract}
\section{Introduction}
\label{sec:intro}

Text-to-video models~\cite{wan} generate high-quality individual shots,
but a short film of eight to sixteen shots across several scenes
additionally requires consistent character appearance, stable spatial
layout and continuous character state across shots. With a black-box
generator, separate requests share no internal state, so a recurring
character may change appearance, the fixed elements of an environment
may shift between viewpoints, and consecutive shots may show
incompatible character states. Repeated textual descriptions constrain
semantic content but not the visual appearance or spatial arrangement
the generator produces.

Existing approaches improve consistency through identity
adapters~\cite{consisid}, reference
attention~\cite{storydiffusion}, camera
control~\cite{cameractrl}, explicit scene representations~\cite{gen3c}
or joint multi-shot generation~\cite{holocine}. All of these
require generator parameters or internal features that a black-box
generator does not expose. Agentic frameworks instead decompose
production into roles and coordinate frozen models around them:
LLM-directed pipelines~\cite{movieagent} plan
scenes and shots with director and screenwriter roles,
AniMaker~\cite{animaker} adds clip selection and review agents,
ViMax~\cite{vimax} coordinates production roles and renders transition
clips, NEWTON~\cite{newton} trains a planner on trajectory-level
rewards, and VISTA~\cite{vista} and UniVA~\cite{univa} add multi-critic
refinement and plan-and-act tool use. In these systems the roles
exchange only text plans, prompts and keyframes, which underdetermine
what the generator renders, and the roles are fixed prompts that never
learn from the shots they produce.

We instead treat condition construction as the interface between
narrative planning and frozen video generation, and realise it in
MVAgent, a multi-agent pipeline whose agents collaborate through typed
conditioning inputs rather than text alone (Fig.~\ref{fig:overview}a),
like the structured outputs of MetaGPT~\cite{metagpt}. Because a single
environment image constrains only the viewpoint it shows,
\emph{generative spatial anchoring} (Sec.~\ref{ssec:anchor}) lets a
Spatial Grounding agent sample views from camera-traversal clips rendered by the
generator and index them by visible fixed elements, so that a view
matching the target camera guides each shot. As generated clips depart
from the planned states, an Observer stores VLM descriptions of each
clip in a \textit{continuity memory} (Sec.~\ref{ssec:dossier}), from which a
Transition agent derives character action and spatial references, so
that each shot opens from the state actually reached. An Orchestrator
composes these inputs into each request. Since a request reveals its
effect only after rendering, we train it by agentic reinforcement
learning~\cite{agenticrl}, the optimization of LLM agents through
multi-step interaction with an environment, here with one costly render
per step and one evaluation per video. Instead of crediting every step
with that evaluation~\cite{agentflow,newton} or relying on states that
recur across rollouts~\cite{gigpo}, \emph{Trunk-GDPO}
(Sec.~\ref{ssec:gdpo}) branches candidates from each shot state,
compares them with group-relative, reward-decoupled
advantages~\cite{dapo,gdpo} and continues the best as the trunk, 
with asynchronous updates~\cite{asyncrlhf} hiding rendering latency.
Our contributions are twofold. First, we design a multi-agent collaboration for agentic video generation that recasts cross-shot consistency as condition construction through generative spatial anchoring and observation-grounded continuity. Second, to train Orchestrator, we propose Trunk-GDPO, which supervises it at shot level within a frozen pipeline and improves consistency and human preference over episode-level training.

\section{Method}
\label{sec:method}

\begin{figure*}[!t]
\centering
\includegraphics[width=0.85\textwidth]{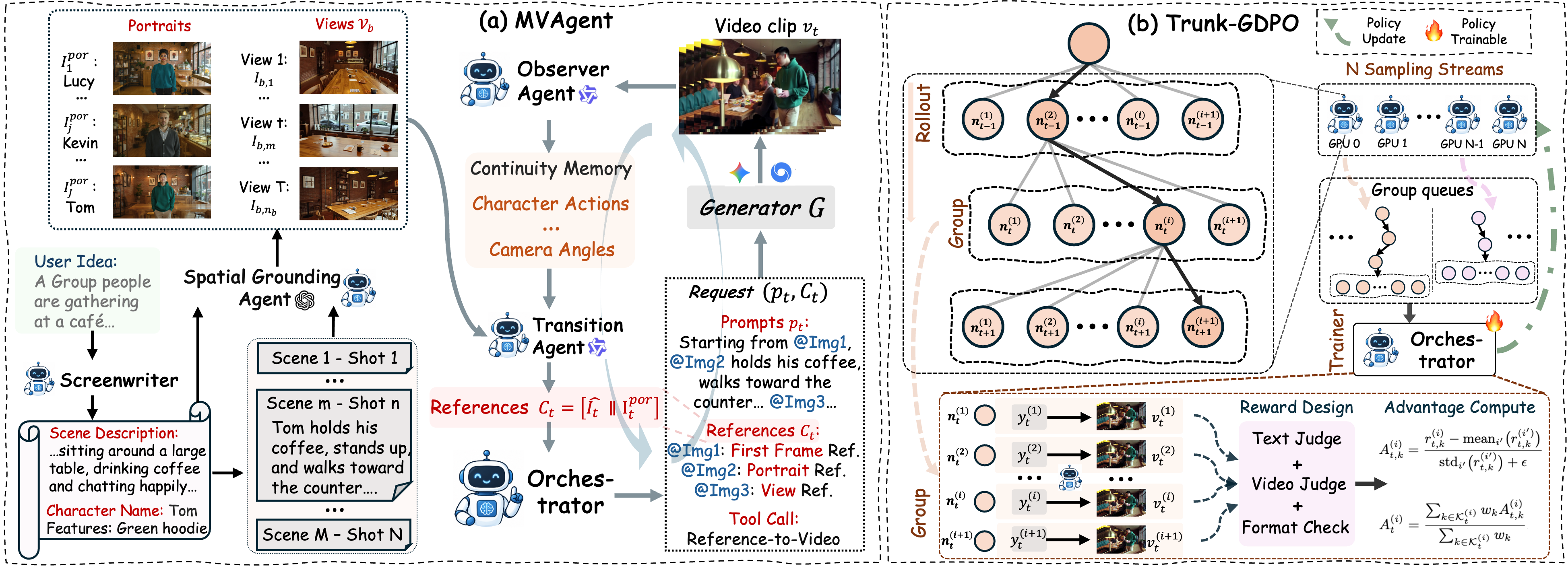}
\vspace{-4pt}
\caption{MVAgent. (a) The Screenwriter and the Spatial Grounding agent
supply the persistent assets, the environment image $I_b$ and the
multi-view library, which anchor character appearance and spatial
layout. The Observer and the Transition agent supply the continuity
memory and the reference $\hat I_t$, which carry character state. The
Orchestrator composes them into each request to the frozen video generator.
(b) Trunk-GDPO trains the Orchestrator. A shot-level candidate group
shares one state and policy version, all candidates are rendered and
judged, components are normalised separately before aggregation, and
the trunk advances the video, with asynchronous sampling and training.}
\label{fig:overview}
\end{figure*}

\textbf{Framework overview.} MVAgent (Fig.~\ref{fig:overview}a) is a
multi-agent pipeline whose agents collaborate through typed conditioning
inputs. The
Screenwriter fixes the persistent assets and the shot plan, the Spatial
Grounding agent supplies a view of the environment for each shot, the
Observer and the Transition agent supply the continuity memory and a
transition reference, and the Orchestrator composes these inputs into
one request for the frozen generator. All roles are frozen except the
Orchestrator's shot-generation policy, which Trunk-GDPO trains with
feedback from frozen judges.

\subsection{Problem statement}
\label{ssec:formulation}

A video consists of $T$ shots produced by a \textit{black-box} video generator $G$ that
is accessed only through its conditioning inputs. Shot $t$ receives a
prompt $p_t$ and an ordered tuple of reference images $\mathcal{C}_t$,
written $[\cdot]$, whose positions carry fixed roles, the first image
anchoring the transition into the shot and the remaining images
identifying the characters,
and yields $v_t\sim G(\cdot\mid p_t,\mathcal{C}_t)$. The Screenwriter
provides the persistent assets, namely a portrait $I^{\mathrm{por}}_j$
of each character and the fixed elements $\mathcal{E}_b$ of each
environment $b$, and a shot plan that specifies for each shot the cast
$\mathcal{P}_t$, the environment $b_t$, a camera specification
$\kappa_t$ (shot size, camera position and facing), the action and
dialogue, and the intended opening and end states. Character appearance
and spatial layout, namely the arrangement of $\mathcal{E}_b$ under
changes of viewpoint, are addressed by the spatial branch
(Sec.~\ref{ssec:anchor}), character state across connected shots by the
continuity branch (Sec.~\ref{ssec:dossier}), and adherence to the shot
plan by the Orchestrator's request composition (Sec.~\ref{ssec:gdpo}).
Since $G$ is fixed and text constrains semantics rather than
appearance, the problem is to construct requests
$\{(p_t,\mathcal{C}_t)\}_{t=1}^{T}$ that carry the persistent assets
together with observations of the generated state.

\subsection{Generative spatial anchoring}
\label{ssec:anchor}

Textual environment descriptions underdetermine spatial layout, so
separate requests place recurring elements differently, and a single
reference image constrains only the viewpoint it shows. The Spatial
Grounding agent therefore anchors each environment $b$ to a multi-view
library that is generated rather than reconstructed. From a
character-free environment image $I_b$ rendered for $\mathcal{E}_b$, it
renders camera-traversal clips
$v^{\mathrm{trav}}_b\sim G(\cdot\mid p^{\mathrm{trav}}_b,[I_b])$,
where the prompt $p^{\mathrm{trav}}_b$ moves the camera through the
environment, so that the temporal coherence of the generator keeps one
layout in view while the viewpoint changes. Frames sampled from these
clips form the candidate set $\mathcal{F}_b$, a candidate joins the view
set $\mathcal{U}_b$ only when it is distinct from every view admitted
before it in sampling order, and the library $\mathcal{V}_b$ pairs each
admitted view with a structured description of its fixed elements and
their arrangement, produced by a VLM $\Phi$:
\begin{equation}
\begin{split}
\mathcal{U}_b&=\{I_b\}\cup\bigl\{f\in\mathcal{F}_b:
\min\nolimits_{u\in\mathcal{U}_b^{\prec f}}\Delta(f,u)\ge\delta\bigr\},\\[-1pt]
\mathcal{V}_b&=\{(I_{b,m},c_{b,m})\}_{m=0}^{n_b},\qquad
c_{b,m}=\Phi(I_{b,m}),
\end{split}
\label{eq:lib}
\end{equation}
where $\mathcal{U}_b^{\prec f}$ denotes the views admitted before $f$,
$\Delta$ is a pixel-space distance with distinctness threshold
$\delta$, and $I_{b,0}=I_b,\dots,I_{b,n_b}$ enumerate $\mathcal{U}_b$.

At generation time the library is queried with the camera
specification: an LLM $\Lambda$ compares $\kappa_t$ with the
descriptions and returns the view whose fixed elements should be
visible under the intended framing,
$m^\star_t=\Lambda(\kappa_t,\{c_{b_t,m}\}_{m})$, with the environment
image as default when no view applies. The retrieved view and its description
condition the transition of Sec.~\ref{ssec:dossier} as a visual
reference and a textual layout constraint, which anchor the arrangement
of fixed elements while leaving framing to the shot plan.

\subsection{Continuity memory}
\label{ssec:dossier}

After each shot the Observer, a VLM, describes the camera configuration
and the position, orientation and action of each visible character at
the end of $v_{t-1}$, and the description is stored in the continuity
memory with the shot plan entry and the request. The Transition agent
turns this memory into a reference for character actions and positions
by rendering a two-shot transition whose first shot restates the
observed state and whose second shot states opening of shot $t$:
\begin{equation}
\begin{split}
\hat I_t&=\operatorname{Ref}\Bigl(G\bigl(\cdot\mid p^{\mathrm{2s}}_t,
[f_{t-1}]\mathbin{\Vert}\mathcal{I}^{\mathrm{por}}_t
\mathbin{\Vert}[I^{\mathrm{sp}}_t]\bigr)\Bigr),\\[-1pt]
\mathcal{C}_t&=[\hat I_t]\mathbin{\Vert}\mathcal{I}^{\mathrm{por}}_t,
\end{split}
\label{eq:cond}
\end{equation}
where $\Vert$ denotes ordered concatenation, $p^{\mathrm{2s}}_t$ is the
prompt of this transition, $f_{t-1}$ is the final frame of $v_{t-1}$,
$\mathcal{I}^{\mathrm{por}}_t=[I^{\mathrm{por}}_j]_{j\in\mathcal{P}_t}$
holds the portraits of the cast, the spatial reference
$I^{\mathrm{sp}}_t$ is the selected view $I_{b_t,m^\star_t}$ when the
environment is unchanged and the environment image $I_{b_t}$ otherwise,
and $\operatorname{Ref}$ extracts the first frame after the cut. The
Spatial Grounding agent verifies that $\hat I_t$ preserves the
arrangement of fixed elements described by $c_{b_t,m^\star_t}$,
irrespective of illumination, and requests one regeneration with a new
seed when the verification fails. The selected continuation updates the
memory for the next shot.

\subsection{Trunk-GDPO}
\label{ssec:gdpo}

The Orchestrator's policy selects a generation mode and writes a prompt
specifying reference usage, planned actions, camera behavior and
applicable continuity constraints. An episode is one video and an agent
step is one request. A terminal reward evaluates a completed video only
once, whereas each rendered shot can be evaluated locally, so Trunk-GDPO
samples $K$ candidates from the same shot state, executes and evaluates
all of them, and uses their relative rewards to supervise the policy,
while one candidate continues the rollout. This candidate is the
\emph{trunk}, by analogy with expanding one branch of a search tree. A
video of $T$ shots yields $T$ comparison groups from $TK$
candidate renders, the same number as $K$ complete videos, and supervises individual decisions under a shared context
without estimating their delayed effects.

\textbf{Groups and advantages.} The state $s_t$ collects the shot plan
entry, the persistent assets, the continuity memory, the selected view
$I_{b_t,m^\star_t}$, the reference tuple $\mathcal{C}_t$ and the
available generation modes $\mathcal{M}_t$. The policy emits a
structured output $y_t$, a JSON object naming the mode and the prompt,
a fixed executor validates it into the request
$a_t=\operatorname{Exec}(s_t,y_t)=(p_t,\mathcal{C}_t)$, and the
feedback $e_t$ comprises the rendered clip $v_t$ and the judge
assessments. With the backends used here
reference-to-video is the only mode, so learning affects prompt
construction. At each shot, $K$ candidates
$\{y^{(i)}_t\}_{i=1}^{K}\sim\pi_{\theta_{\mathrm{old}}}(\cdot\mid s_t)$
share one state and one behavior-policy version and receive component
rewards $r^{(i)}_{t,k}=R_k(s_t,y^{(i)}_t,e^{(i)}_t)$ for
$k\in\mathcal{K}$. Because the components differ in scale
and applicability, each available component is normalised across the
candidates before aggregation, following GDPO~\cite{gdpo}:
\begin{equation}
A^{(i)}_{t,k}=\frac{r^{(i)}_{t,k}-\mean_{i'}\bigl(r^{(i')}_{t,k}\bigr)}
{\std_{i'}\bigl(r^{(i')}_{t,k}\bigr)+\epsilon},\qquad
A^{(i)}_t=\frac{\sum_{k\in\mathcal{K}^{(i)}_t} w_k A^{(i)}_{t,k}}
{\sum_{k\in\mathcal{K}^{(i)}_t} w_k},
\label{eq:adv}
\end{equation}
where $\mathcal{K}^{(i)}_t$ contains the components available for
candidate $i$, tied components yield zero advantage, and no value
function is learned. The candidate with the highest combined advantage
advances shot-level video generation and updates the memory,
\begin{equation}
s_{t+1}=\mathcal{T}\bigl(s_t,\,y^{(i^\star_t)}_t,\,e^{(i^\star_t)}_t\bigr),\qquad
i^\star_t=\argmax_{i}\,A^{(i)}_t.
\label{eq:trunk}
\end{equation}
 
\textbf{Objective and LoRA policy.} Let $\rho^{(i)}_{t,j}$ denote the
current-to-behavior probability ratio for token $j$ of candidate $i$.
The policy maximises a clipped surrogate~\cite{ppo} regularised towards
the frozen reference policy $\pi_{\mathrm{ref}}$:
\begin{equation}
\begin{split}
\mathcal{J}(\theta)=&\mathbb{E}\Bigl[\frac{1}{\sum_{i}|y^{(i)}_t|}
\sum_{i=1}^{K}\sum_{j=1}^{|y^{(i)}_t|}\min\bigl\{\rho^{(i)}_{t,j}A^{(i)}_t,\\[-4pt]
&\ \clip(\rho^{(i)}_{t,j},1{-}\varepsilon_{\ell},1{+}\varepsilon_{h})A^{(i)}_t\bigr\}\Bigr] -\beta\,\mathbb{D}_{\mathrm{KL}}\bigl(\pi_\theta\,\|\,\pi_{\mathrm{ref}}\bigr),
\end{split}
\label{eq:objective}
\end{equation}
where the asymmetric clipping follows DAPO~\cite{dapo}, response tokens
are averaged within the group, each group supports one optimizer update,
and the KL penalty uses the per-token $k_3$ estimator.
Training updates low-rank adapters~\cite{lora} on the attention
projections with zero initial effective update, so the initial policy is the base model
and disabling the adapters recovers $\pi_{\mathrm{ref}}$, and only
adapter parameters move between the trainer and the sampling workers.

\textbf{Asynchronous training.} To overlap rendering with optimization,
$N$ sampling streams generate videos and enqueue each completed group,
and one trainer performs one update per group and publishes adapter
parameters every $B$ updates (Fig.~\ref{fig:overview}b). Streams reload
parameters only at shot boundaries, so the $K$ candidates of a group
share one behavior policy while different shots of a video may use
different versions. Groups older than $\sigma_{\max}$ versions are
discarded, so the policy staleness of every update is bounded, and
stored token IDs and behavior log-probabilities let the trainer
recompute current-policy probabilities in the same contexts, which keeps
the ratios consistent with the stored samples.

\textbf{Reward components.} The components in Eq.~\eqref{eq:adv} are a
format, a text and a video reward,
$k\in\mathcal{K}=\{\mathrm{fmt},\mathrm{txt},\mathrm{vid}\}$, whose
advantages are weighted by $w_{\mathrm{fmt}}=0.15$,
$w_{\mathrm{txt}}=0.35$ and $w_{\mathrm{vid}}=0.50$. The format reward
$r^{(i)}_{t,\mathrm{fmt}}$ is 1 if the executor accepts $y^{(i)}_t$ and
0 otherwise, and weighs least because a larger weight rewarded
well-formed but empty prompts. The text reward
$r^{(i)}_{t,\mathrm{txt}}$ is one fifth of the text judge's mean 1--5
rating on criteria including plan adherence, visual specificity and
reference usage. For the video reward $r^{(i)}_{t,\mathrm{vid}}$, the
video judge ranks the $K$ clips on acting (act), physical plausibility
(phy) and camera work (cam), which lack a reference, but checks
consistency (cons) against the portraits and the selected view, since
ranking would still reward the best of a failing group:
\begin{equation}
\begin{split}
r^{(i)}_{t,\mathrm{vid}}&=\textstyle\sum_{d\in\mathcal{D}}\lambda_d\,r^{(i)}_{t,d},\qquad
\mathcal{D}=\{\mathrm{act},\mathrm{phy},\mathrm{cam},\mathrm{cons}\},\\
r^{(i)}_{t,d}&=\begin{cases}
1-\mathrm{rk}^{(i)}_{t,d}/(K-1), & d\in\{\mathrm{act},\mathrm{phy},\mathrm{cam}\},\\
q^{(i)}_{t,+}/\bigl(q^{(i)}_{t,+}+q^{(i)}_{t,-}\bigr), & d=\mathrm{cons},
\end{cases}
\end{split}
\label{eq:vid}
\end{equation}
where $\lambda_d$ are fixed weights summing to one,
$\mathrm{rk}^{(i)}_{t,d}$ is the zero-based rank of clip $i$, and
$q^{(i)}_{t,+}$ and $q^{(i)}_{t,-}$ count the appearance attributes and
fixed elements the clip matches and contradicts.

\begin{table*}[!t]
\centering\footnotesize
\caption{Cross-shot consistency on ViMax-Bench (ViCLIP cosine similarity
for cross-scene CC, intra-scene IC and global GC, higher is better). Agentic rows name the LLM and the generator, bold and underline
mark the highest and second-highest scores.}
\label{tab:cons}
\setlength{\tabcolsep}{4pt}
\begin{tabular}{@{}l ccc ccc ccc@{}}
\toprule
\multirow{2}{*}{Method}
  & \multicolumn{3}{c}{Medium} & \multicolumn{3}{c}{Long}
  & \multicolumn{3}{c}{Overall} \\
\cmidrule(lr){2-4}\cmidrule(lr){5-7}\cmidrule(l){8-10}
  & CC & IC & GC & CC & IC & GC & CC & IC & GC \\
\midrule
Veo 3.1                     & 0.5536 & 0.5640 & 0.5582 & 0.4687 & 0.4778 & 0.4704 & 0.4978 & 0.5074 & 0.5005 \\
IC-LoRA                     & 0.5398 & 0.5713 & 0.5537 & 0.5015 & 0.5328 & 0.5087 & 0.5147 & 0.5460 & 0.5241 \\
StoryDiffusion              & 0.5606 & 0.5796 & 0.5691 & 0.5007 & 0.5195 & 0.5048 & 0.5212 & 0.5401 & 0.5269 \\
HoloCine                    & 0.5366 & 0.5892 & 0.5597 & 0.4940 & 0.5565 & 0.5080 & 0.5098 & 0.5686 & 0.5272 \\
ViMax (Gemini 3.1 Flash, Veo 3.1)   & 0.5681 & 0.5914 & 0.5785 & 0.5088 & 0.5402 & 0.5157 & 0.5291 & 0.5577 & 0.5372 \\
ViMax (Gemini 3 Pro, Veo 3.1)       & 0.5886 & \textbf{0.6046} & \underline{0.5957} & \underline{0.5298} & \underline{0.5629} & \underline{0.5369} & \underline{0.5500} & \underline{0.5772} & \underline{0.5571} \\
MVAgent (Gemini 3.1 Flash, Veo 3.1) & \underline{0.5926} & 0.6012 & 0.5890 & 0.5126 & 0.5494 & 0.5271 & 0.5403 & 0.5602 & 0.5450 \\
\textbf{MVAgent (Gemini 3 Pro, Veo 3.1)} & \textbf{0.5997} & \underline{0.6029} & \textbf{0.5978} & \textbf{0.5348} & \textbf{0.5704} & \textbf{0.5437} & \textbf{0.5611} & \textbf{0.5863} & \textbf{0.5689} \\
\bottomrule
\end{tabular}
\end{table*}

\section{Experiments}
\label{sec:exp}

\textbf{Setup.} Each final shot uses one policy output at inference,
without best-of-$K$ selection or post-generation repair. Unless a table row names another LLM,
the LLM-based agents use GPT-5.4, the VLM-based agents use Gemini~3~Pro
and the video generator is Veo~3.1. For Trunk-GDPO training only the LLM
in the Orchestrator role is replaced by the open-source
Qwen3-32B~\cite{qwen3} with rank-64 LoRA ($\alpha=128$), and
Table~\ref{tab:abl} uses this
configuration. The frozen text and video judges are qwen3.8-max and
Gemini~3.1~Pro. Training uses $K=4$,
$(\varepsilon_\ell,\varepsilon_h)=(0.2,0.3)$, $\beta=10^{-3}$, learning
rate $10^{-5}$, $B=4$, $\sigma_{\max}=5$, $N=4$ sampling streams on four
GPUs, one trainer on four GPUs.

\textbf{Benchmarks and metrics.} ViMax-Bench~\cite{vimax} contains 35
stories (12 Medium with two scenes and 8--10 shots, 23 Long with three to
four scenes and 12--16 shots) covering character persistence, background
persistence and multi-person interaction. Following its protocol, we
encode each shot with ViCLIP~\cite{viclip} and report intra-scene (IC),
cross-scene (CC) and global (GC) cosine similarity. Narrative quality
follows the rubric of~\cite{vimax} on 50 NarrativeQA
novels~\cite{narrativeqa}, where GPT-5.4 scores character behavior
consistency (CBC), narrative coherence (NC), plot pacing and rhythm
(PPR), visual specificity and shot usability (VSU), scene transition
quality (STQ) and faithfulness (NF) on a 1--5 scale. In the pairwise
protocol of~\cite{vimax}, three raters compare video pairs shown in
random order on cross-scene consistency (CSC), semantic following (SF)
and aesthetic quality (AQ), reported as preference rates in percent.

\textbf{Cross-shot consistency.} Table~\ref{tab:cons} reports results
alongside the baseline scores from~\cite{vimax}. With Gemini~3~Pro and
Veo~3.1, MVAgent achieves the highest score in eight of nine columns,
including overall GC (0.5689 versus 0.5571 for ViMax), and with
Gemini~3.1~Flash and Veo~3.1 it exceeds ViMax in every column, both
under the same language and video backends.

\begin{figure}[t]
  \centering
  \includegraphics[width=\linewidth]{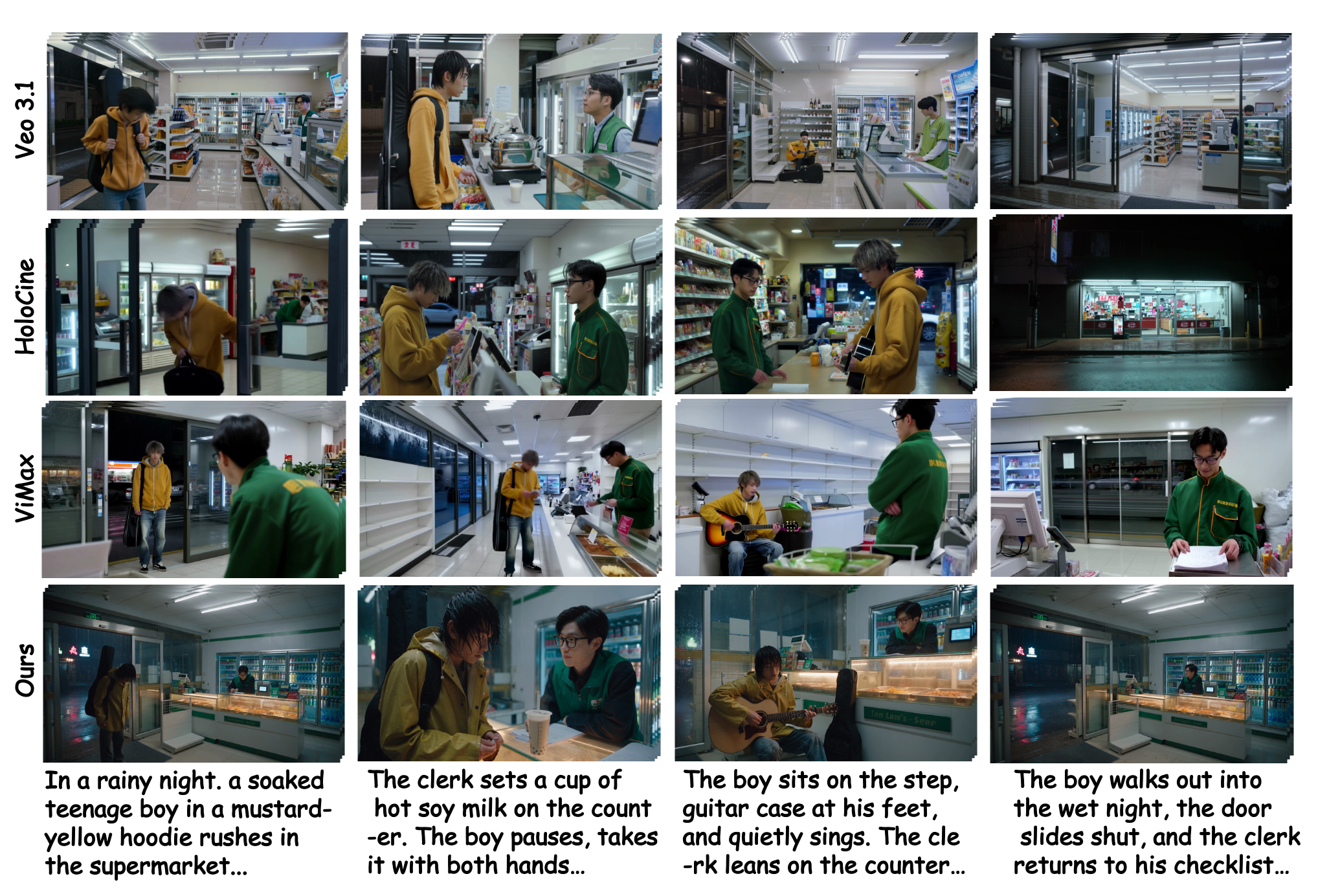}
  \vspace{-16pt}
  \caption{Qualitative comparison of three conditioning strategies for a
  multi-shot story. Each row shows keyframes generated with the same video
  model, story and camera specifications.}
  \label{fig:comparison}
\end{figure}

\textbf{Qualitative comparison.} Fig.~\ref{fig:comparison} compares three
conditioning strategies with the generator, story and camera
specifications held fixed. MVAgent preserves character appearance
across the four shots, and the store
retains its arrangement of doors, counters and shelves under different
framings. Text-only conditioning changes the counter, freezer and door
arrangement and the customer's appearance despite repeated descriptions,
and inconsistent descriptions produce larger changes in both.

\begin{table}[!t]
\centering\footnotesize
\caption{Narrative-planning quality (1--5 scale) with different frozen agent
backbones, following novel-to-video protocol of~\cite{vimax}.}
\label{tab:backbone}
\setlength{\tabcolsep}{3pt}
\begin{tabular}{@{}lccccccc@{}}
\toprule
Backbone & CBC & NC & PPR & VSU & STQ & NF & Avg. \\
\midrule
Ours-Gemini 3 Pro      & \underline{4.37} & \underline{4.08} & {4.19} & \underline{4.83} & 4.01 & \underline{2.64} & \underline{4.02} \\
Ours-Gemini 3.1 Flash  & 4.29 & 3.96 & \underline{4.20} & 4.76 & \underline{4.10} & {2.57} & 3.98 \\
Ours-GPT-5.4           & \textbf{4.54} & \textbf{4.22} & \textbf{4.41} & \textbf{4.90} & \textbf{4.23} & \textbf{2.82} & \textbf{4.19} \\
\bottomrule
\end{tabular}
\end{table}

\textbf{Narrative quality and LLM backbone.} MVAgent achieves the highest
mean score in the novel-to-video comparison (4.256 versus 4.107 for the
strongest agentic baseline). Its advantage does not extend to
faithfulness, which is consistent with the absence of retrieval over the
novel. Table~\ref{tab:backbone} compares frozen agent backbones with the
other roles fixed. GPT-5.4 obtains the highest score on all six
dimensions with an average of 4.19. Faithfulness is the lowest dimension for
every backbone and visual specificity the highest. 

\begin{table}[!t]
\centering\footnotesize
\caption{Pairwise human preference rates (\%) for MVAgent against each
baseline on ViMax-Bench.}
\label{tab:human}
\setlength{\tabcolsep}{5pt}
\begin{tabular}{@{}lcccc@{}}
\toprule
MVAgent vs. & CSC & SF & AQ & Avg. \\
\midrule
Veo 3.1        & 78.85\% & 78.38\% & 71.06\% & 76.07\% \\
StoryDiffusion & 82.79\% & 79.62\% & 77.94\% & 80.12\% \\
HoloCine       & 65.31\% & 68.11\% & 57.45\% & 63.49\% \\
ViMax          & 61.39\% & 63.02\% & 55.78\% & 60.06\% \\
\bottomrule
\end{tabular}
\end{table}

\textbf{Subjective evaluation.} Table~\ref{tab:human} reports pairwise
human preferences against four baselines. MVAgent's preference rate
exceeds 50\% on every criterion in each comparison, with averages from
60.06\% against ViMax to 80.12\% against StoryDiffusion. Against ViMax
the rates are 61.39\% for cross-scene consistency, 63.02\% for semantic
following and 55.78\% for aesthetic quality, so the larger margins
concern consistency and semantic following.

\begin{table}[!t]
\centering\footnotesize
\caption{Ablation on ViMax-Bench with consistency as in
Table~\ref{tab:cons} and preference rates (\%) of the full model against
each variant under the protocol of Table~\ref{tab:human}.}
\label{tab:abl}
\setlength{\tabcolsep}{2.6pt}
\begin{tabular}{@{}l ccc cccc@{}}
\toprule
\multirow{2}{*}{Variant} & \multicolumn{3}{c}{Consistency} & \multicolumn{4}{c}{Subjective evaluation (\%)} \\
\cmidrule(lr){2-4}\cmidrule(l){5-8}
 & CC & IC & GC & CSC & SF & AQ & Avg. \\
\midrule
Portraits only & 0.5286 & 0.5597 & 0.5360 & 71.5 & 56.3 & 53.9 & 60.6 \\
w/o multi-view library & 0.5319 & 0.5624 & 0.5388 & 69.1 & 55.8 & 52.7 & 59.2 \\
Planned state only & 0.5382 & 0.5690 & 0.5417 & 67.2 & 54.3 & 52.9 & 58.1 \\
\midrule
Ours (w/o RL) & 0.5239 & 0.5496 & 0.5204 & 63.6 & 54.4 & 51.1 & 56.3 \\
Episode-level RL & 0.5362 & 0.5681 & 0.5403 & 55.0 & 52.9 & 51.2 & 53.0 \\
\textbf{Ours} & 0.5417 & 0.5712 & 0.5438 & - & - & - & - \\
\bottomrule
\end{tabular}
\end{table}

\textbf{Ablation.} Table~\ref{tab:abl} evaluates conditioning components
and training configurations. The conditioning variants use portraits
alone, retain the environment image without the multi-view library, or
replace observed character states with planned states. The full model
achieves the highest CC, IC and GC and is preferred over every variant, from 60.6\% against
portraits alone to 58.1\% against planned states, so the margin shrinks
as components are added. The training comparison contrasts the untrained
base policy, episode-level RL, which groups complete-video rollouts under
a terminal reward per video, and Trunk-GDPO, which forms candidate groups
at individual shot states. The full model is preferred over the untrained policy in 56.3\% and
over episode-level RL in 53.0\% of comparisons, chiefly on cross-scene
consistency (63.6\% and 55.0\%). Within each group, both metrics agree
on the ranking.

\section{Conclusion}
\label{sec:concl}

MVAgent formulates multi-shot narrative video generation as condition
construction for a frozen generator, carried out by agents that exchange
typed conditioning inputs. A Spatial Grounding agent supplies
reusable spatial references, an Observer and a Transition agent turn VLM
observations into continuity memory and action and position references,
and Trunk-GDPO trains the Orchestrator that composes them into requests
with shot-level feedback, subject to the generator's capabilities and
the judges' reliability.

\vfill\pagebreak

\bibliographystyle{IEEEbib}
\bibliography{strings,refs}

\end{document}